\documentclass[lettersize,journal]{IEEEtran}
\usepackage{amsmath,amsfonts}
\usepackage{algorithmic}
\usepackage{algorithm}
\usepackage{array}
\usepackage[caption=false,font=normalsize,labelfont=sf,textfont=sf]{subfig}
\usepackage{textcomp}
\usepackage{stfloats}
\usepackage{url}
\usepackage{verbatim}
\usepackage{graphicx}
\usepackage{cite}
\usepackage{multirow}  
\usepackage{booktabs}  
\usepackage{url}
\usepackage[breaklinks,colorlinks,allcolors=blue]{hyperref}

\usepackage{xcolor}	
\usepackage{colortbl}	
\usepackage{cleveref} 

\begin{document}

\title{AeroDeshadow: Physics-Guided Shadow Synthesis and Penumbra-Aware Deshadowing for Aerospace Imagery}

\author{Wei Lu, Zi-Yang Bo, Fei-Fei Sang, Yi Liu, Xue Yang, and Si-Bao Chen$^\ast$
	\thanks{This work was supported in part by the NSFC Key Project of Joint Fund for Enterprise Innovation and Development under Grant U24A20342, and in part by the National Natural Science Foundation of China under Grant 62576006 and 61976004. (Corresponding author: Si-Bao Chen)}
	\thanks{Wei Lu, Zi-Yang Bo, Fei-Fei Sang, Yi Liu, and Si-Bao Chen are with the MOE Key Laboratory of ICSP, IMIS Laboratory of Anhui, Anhui Provincial Key Laboratory of Multimodal Cognitive Computation, Zenmorn-AHU AI Joint Laboratory, School of Computer Science and Technology, Anhui University, Hefei 230601, China (e-mail:luwei\_ahu@qq.com; e24201058@stu.ahu.edu.cn; ffeisang@163.com; e125211073@stu.ahu.edu.cn; sbchen@ahu.edu.cn).}
	\thanks{Xue Yang is with the School of Automation and Intelligent Sensing, Shanghai Jiao Tong University, Shanghai 200240, China (e-mail: yangxue-2019-sjtu@sjtu.edu.cn).}
}

\markboth{IEEE TRANSACTIONS ON IMAGE PROCESSING, 2026}%
{Shell \MakeLowercase{\textit{et al.}}: A Sample Article Using IEEEtran.cls for IEEE Journals}

\maketitle

\begin{abstract}
	Shadows are prevalent in high-resolution aerospace imagery (ASI). They often cause spectral distortion and information loss, which degrade downstream interpretation tasks. While deep learning methods have advanced natural-image shadow removal, their direct application to ASI faces two primary challenges. First, strictly paired training data are severely lacking. Second, homogeneous shadow assumptions fail to handle the broad penumbra transition zones inherent in aerospace scenes. To address these issues, we propose AeroDeshadow, a unified two-stage framework integrating physics-guided shadow synthesis and penumbra-aware restoration. In the first stage, a Physics-aware Degradation Shadow Synthesis Network (PDSS-Net) explicitly models illumination decay and spatial attenuation. This process constructs AeroDS-Syn, a large-scale paired dataset featuring soft boundary transitions. Constrained by this physical formulation, a Penumbra-aware Cascaded DeShadowing Network (PCDS-Net) then decouples the input into umbra and penumbra components. By restoring these regions progressively, PCDS-Net alleviates boundary artifacts and over-correction. Trained solely on the synthetic AeroDS-Syn, the network generalizes to real-world ASI without requiring paired real annotations. Experimental results indicate that AeroDeshadow achieves state-of-the-art quantitative accuracy and visual fidelity across synthetic and real-world datasets. The datasets and code will be made publicly available at: \url{https://github.com/AeroVILab-AHU/AeroDeshadow}.
\end{abstract}


\section{Introduction}\label{Introduction}
\IEEEPARstart{H}{igh-resolution} optical aerospace imagery (ASI) is widely used in Earth observation tasks, including urban planning \cite{zhu2017deep,cheng2017remote,lu2023robust} and disaster monitoring \cite{wang2024attention, shu2026uniroute}. However, shadows cast by elevated ground objects \cite{arevalo2008shadow} and specific atmospheric conditions \cite{zhu2025real} are prevalent. These shadows introduce severe illumination inconsistency, leading to the loss of texture details, spectral distortion, and ambiguity in object boundaries \cite{lu2025legnet,10990319,11202372,lu2026unravelnet}. Consequently, the performance of downstream tasks \cite{xia2018dota,lu2024decouplenet,lu2026lwganet} is often degraded. Therefore, effective shadow removal is essential for the reliable ASI interpretation.

Existing shadow removal methods are broadly categorized into physical-model-based, statistical, and deep learning approaches. Traditional methods often rely on simplified illumination assumptions or statistical feature matching \cite{finlayson2006removal,9601181,guo2012paired,murali2013shadow}, rendering them inadequate for modeling spatially variant shadow patterns in complex aerospace scenes. Recently, deep learning has improved restoration quality on standard benchmarks in natural scenes \cite{qu2017deshadownet, wang2018stacked}. Nevertheless, the success of these methods usually relies on two assumptions: the availability of large-scale paired shadow/shadow-free datasets, and the treatment of shadow regions as homogeneous areas represented by binary masks \cite{chen2021canet, guo2023shadowdiffusion}. Both assumptions are difficult to satisfy in ASI, revealing two fundamental limitations of existing methods.

First, strictly paired training data are rarely available. Satellite revisiting cycles, aerial flight trajectories, and dynamic environmental changes make it impractical to capture the exact same scene with and without shadows under consistent conditions \cite{zhu2017deep}. Existing attempts to alleviate this, such as UAV-based acquisition \cite{luo2022uavsc}, often fail to maintain consistent illumination and imaging conditions, while virtual-scene rendering \cite{chu2024gta} usually exhibits a substantial domain gap with real ASI. 
Although unsupervised and semi-supervised methods \cite{guo2023boundary,UPShadowGAN, liu2021shadow, lu2025unsupervised,jin2021dc} relax the paired-data requirement, the absence of explicit pixel-level supervision still limits the recovery of fine structures and complex textures.

\begin{figure*}[t]
\includegraphics[width=1\linewidth]{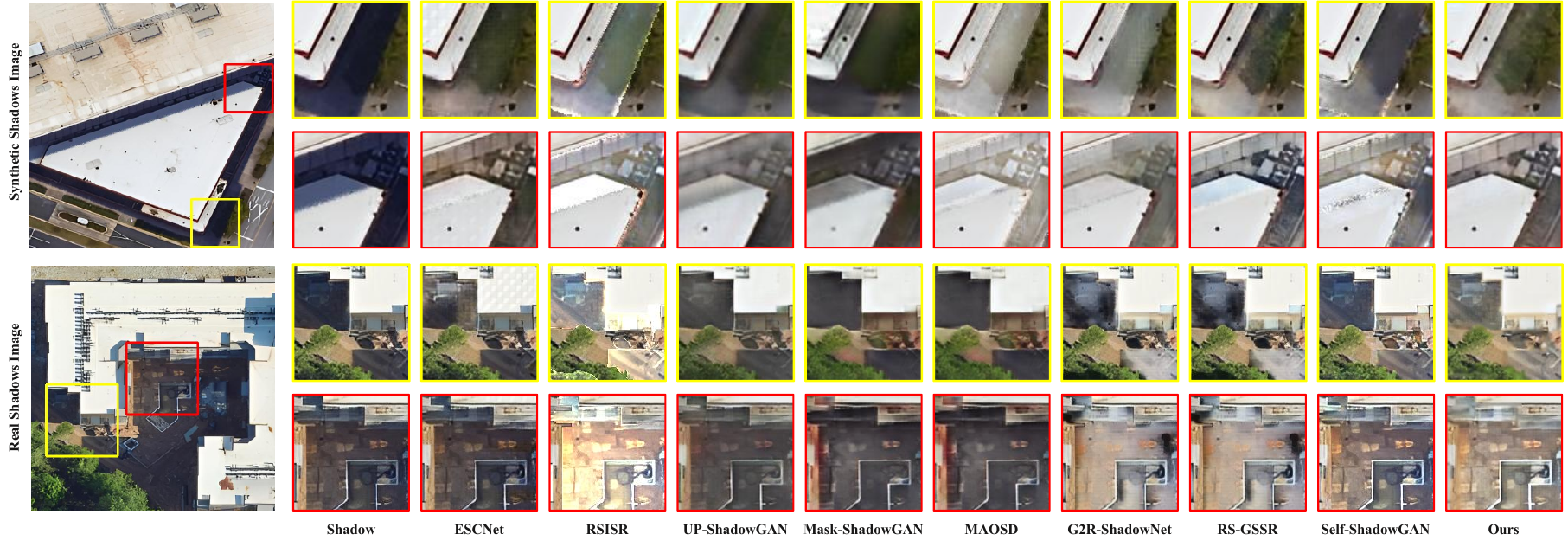}\vspace{-1mm}
\caption{Visual comparison of shadow removal on synthetic (top) and real-world (bottom) aerospace imagery from the proposed AeroDS dataset. The boxes highlight regions with complex color variations and gradual illumination transitions. Compared to competing methods, the proposed approach produces fewer visible artifacts at shadow boundaries and maintains more consistent color in penumbra regions, as highlighted in the red and yellow boxes.} \label{fig:abstract}
\end{figure*}

Second, shadows in ASI exhibit significantly different characteristics from those in natural images. Due to complex scene geometry, long-distance illumination propagation, and atmospheric scattering, ASI shadows usually consist of a dark umbra region surrounded by a broad penumbra transition zone, where illumination and spectral responses vary continuously. Most existing methods, however, formulate shadow removal as a binary restoration problem and treat the entire shadow area as a homogeneous region under a hard mask. Directly applying this uniform restoration tends to over-enhance the umbra while under-restoring the penumbra, or vice versa. As a result, the restored images frequently exhibit blurred structures, spectral inconsistency, and visible artifacts around shadow boundaries, as shown in Fig. \ref{fig:abstract}. These observations indicate that effective ASI shadow removal requires explicit decomposition of the shadow into umbra and penumbra components, followed by progressive restoration of the two regions.

The mutual dependence of these two limitations poses a compounding challenge. The lack of physically meaningful synthetic data limits the supervision of penumbra dynamics, and conversely, inadequate penumbra-aware architectures bottleneck the potential of high-quality data.
		
\begin{figure}[t]
	\includegraphics[width=0.9\linewidth]{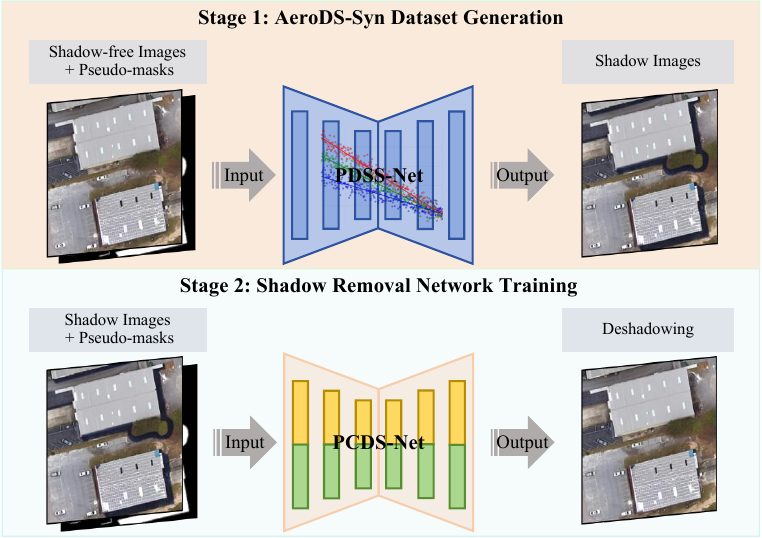}\vspace{-1mm}
	\caption{Overall architecture of the proposed AeroDeshadow framework. The framework bridges the synthetic-to-real domain gap via two stages. Stage 1 employs Physics-aware Degradation Shadow Synthesis Network (PDSS-Net) to construct a large-scale paired dataset (AeroDS-Syn) by simulating realistic shadow attenuation. Stage 2 utilizes Penumbra-aware Cascaded DeShadowing Network (PCDS-Net), trained on AeroDS-Syn, to perform progressive, decoupling-based shadow restoration on real-world aerospace imagery.} \label{fig_2} 
\end{figure}

To address these challenges, we propose AeroDeshadow, a unified two-stage framework integrating physics-guided shadow synthesis and penumbra-aware shadow removal, as illustrated in Fig. \ref{fig_2}.
		
To overcome the paired data bottleneck, we first develop a Physics-aware Degradation Shadow Synthesis Network (PDSS-Net). Unlike virtual-scene rendering or simple intensity attenuation, PDSS-Net introduces a physically interpretable illumination degradation model together with a spatial attenuation mechanism. By applying these constraints to real shadow-free ASI under the guidance of pseudo-shadow masks, PDSS-Net generates realistic shadows with soft boundary transitions, mitigating the domain gap directly at the data source level. Through this pipeline, we construct a large-scale paired synthetic dataset, AeroDS-Syn (2,000 for training, 260 for testing), alongside a dedicated real-scene test set, AeroDS-Real (260 natural shadow samples), for out-of-distribution generalization evaluation.

Based on AeroDS-Syn, we further design a Penumbra-aware Cascaded DeShadowing Network (PCDS-Net). Breaking from the conventional homogeneous shadow assumption, PCDS-Net explicitly decouples the shadow into umbra and penumbra components. Since the recovery of the penumbra depends on accurate illumination compensation within the umbra, PCDS-Net restores these two regions progressively in a cascaded manner. This structural design effectively prevents over-correction near the shadow center and suppresses residual artifacts around the boundaries. Trained solely on AeroDS-Syn, PCDS-Net generalizes robustly to real-world ASI without requiring paired real annotations.

Ultimately, our decoupled, physics-guided formulation seeks to advance beyond conventional uniform mappings by explicitly capturing the spatial heterogeneity between the umbra and penumbra. Specifically, the main contributions of this work are summarized as follows:
\begin{enumerate}
	\item We formulate ASI shadow formation as a spatially heterogeneous physical process and propose AeroDeshadow, a unified two-stage framework. By coupling physics-guided shadow synthesis with penumbra-aware restoration, this provides a principled way to bridge the synthetic-to-real domain gap in aerospace imagery.
	\item To instantiate the proposed physics-decoupled modeling, we design PDSS-Net to simulate illumination degradation and non-linear spatial attenuation. PDSS-Net generates paired training data with physically consistent soft boundaries, enabling the construction of the large-scale AeroDS-Syn dataset.
	\item To realize the progressive restoration mechanism, we develop PCDS-Netto explicitly separate shadow regions into umbra and penumbra components. By reconstructing these components according to their distinct degradation dynamics, the network effectively mitigates boundary artifacts and local over-correction.
	\item We establish a comprehensive benchmark comprising AeroDS-Syn and AeroDS-Real to evaluate shadow removal in ASI. Extensive experiments indicate that the proposed framework achieves state-of-the-art (SOTA) performance in both quantitative and qualitative results.
\end{enumerate}

\section{Related Work}	\label{Related}

\subsection{Shadow Synthesis}	
Early research on shadow synthesis primarily focused on rendering techniques based on computer graphics. For instance, classical methods simulate ray tracing through layered depth images or occluder buffers to achieve soft shadow generation \cite{1407859}. However, these traditional approaches highly depend on scene geometric priors and 3D surface information. Under complex material interactions, they incur significant computational overhead and are prone to visual aliasing, making it difficult to directly simulate the complex light and shadow evolution in ASI.

In recent years, deep learning has advanced shadow synthesis from a data-driven perspective. To alleviate the reliance on strictly paired datasets, Generative Adversarial Networks (GANs)—such as CycleGAN \cite{CycleGAN}, Mask-ShadowGAN \cite{hu2019mask}, and G2R-ShadowNet \cite{G2R_ShadowNet}—utilize unpaired domain translation, cycle-consistency loss, or mask-guided style transfer to enhance the realism of synthetic shadows. Additionally, diffusion models have been introduced to guide the generation process using geometric priors \cite{11094893}. Concurrently, exposure fusion methods demonstrate promising visual results by fusing multiple under-exposed images at the pixel level to synthesize soft shadows \cite{9578628, 10043015}.

However, most of these studies focus on natural imagery. Directly transferring them to the ASI domain often yields uniformly dark shadows with hard edges, failing to simulate the gradual spatial attenuation characteristic of real aerospace scenes. While a few recent works attempt to incorporate illumination degradation priors for shadow synthesis in remote sensing data \cite{10641639, RS_GSSR}, explicitly modeling the spatial transition of the penumbra region remains underexplored.

\subsection{Shadow Removal}
Traditional shadow removal methods mainly rely on physical priors and statistical modeling. Early studies focused on the physical mechanisms of shadow formation, separating luminance and chrominance components or utilizing paired region matching to restore the illumination of shadowed areas \cite{finlayson2006removal, guo2012paired}. Other representative methods employ illumination recovering optimization \cite{7180373} or physics-based image decomposition \cite{9601181} for detail compensation. Although these methods possess a degree of physical interpretability, their performance highly depends on strong prior assumptions or manual intervention, limiting their generalization capability across large-scale, heterogeneous ASI scenes.

With the public release of benchmark datasets (e.g., SRD \cite{qu2017deshadownet}, ISTD \cite{wang2018stacked}, and WSRD \cite{10208694}), deep learning has demonstrated capacity in shadow removal. Transformer-based methods utilize their long-range modeling capabilities to capture cross-region correlations, achieving refined removal in complex scenes \cite{Guo_Huang_Liu_Cheng_Wen_2023, 10658435, 10677878}. GAN-based and CNN-based methods continuously approximate the probability distribution of shadow-free images through adversarial training and context-aware feature extraction \cite{chen2021canet, jin2021dc, Jiang2023}. Concurrently, generative diffusion models utilize powerful prior distributions and iterative denoising to achieve high-fidelity shadow elimination in the latent space \cite{guo2023shadowdiffusion, 10483579, Diff-Shadow_2025,Zeng2025 }. Despite their success in natural images, these methods typically formulate shadow removal as a binary mapping guided by a hard mask, treating the entire shadow as a homogeneous region. This formulation ignores the gradual illumination transition of the penumbra, frequently leading to noticeable boundary artifacts when applied to ASI.

In the aerospace domain, the removal of complex surface and cloud shadows has become a critical research hotspot. Several evolutionary networks and Mamba-based architectures have been proposed to achieve efficient global feature reconstruction under large-scale ASI shadows \cite{luo2022uavsc, chu2024gta, 10833852}. Furthermore, to address the paired data scarcity, researchers have explored unpaired shadow removal techniques by modeling appearance recreation or lightness guidance \cite{liu2021shadow, UPShadowGAN}. However, lacking explicit pixel-level supervision, these unpaired methods are inherently limited in fully reconstructing fine textures. How to effectively integrate the physical priors of shadow formation with deep feature representations, while effectively handling the distinct penumbra transitions in ASI without relying on real paired annotations, remains a critical challenge to be overcome.

\section{Dataset} \label{dataset}
In natural-image shadow removal, paired training samples are typically acquired by removing physical occluders or simulating controlled illumination in virtual environments. In contrast, for real-world ASI, constrained by satellite revisit cycles, aerial flight trajectories, and dynamic atmospheric conditions, it is nearly impossible to acquire pixel-level paired images of the same scene with and without shadows. Therefore, constructing realistic paired training data requires revisiting the physical formation mechanism of shadows.

Photometrically, surface irradiance is mainly composed of direct sunlight and diffuse skylight. Cast shadows are produced when direct illumination is partially or completely occluded, leaving only diffuse illumination. This process constitutes a physically meaningful degradation from shadow-free appearance to shadowed appearance. This prior provides the basis for learning realistic shadow distributions from unpaired ASI and generating paired training samples.

\begin{figure}[t]
	\includegraphics[width=1\linewidth]{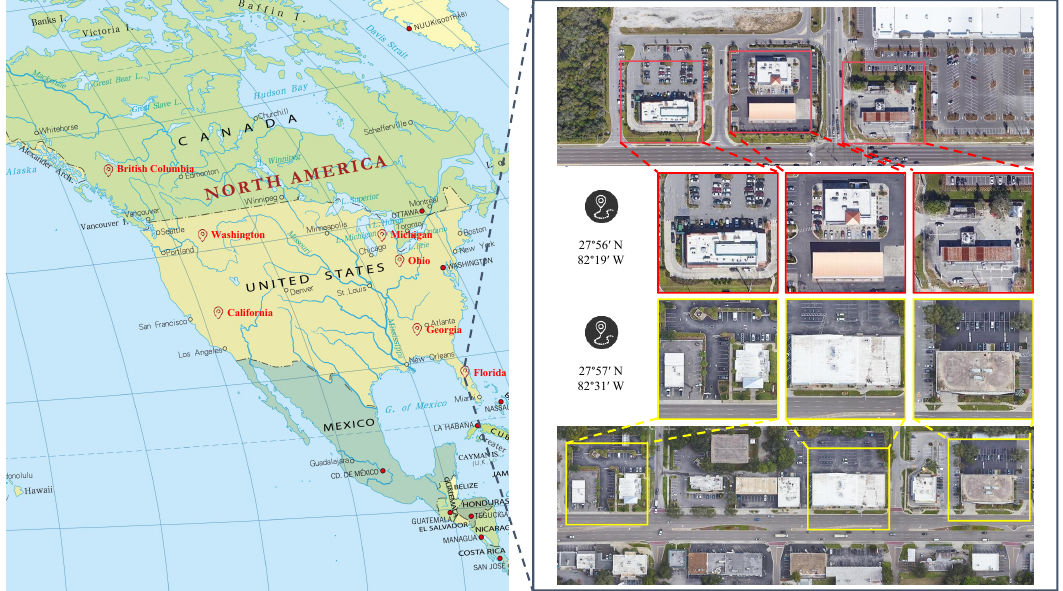}
	\caption{Geographical distribution and representative samples of AeroDS. Multi-temporal shadowed and shadow-free ASI are collected from diverse regions worldwide, providing realistic samples for learning the distribution of ASI shadows.}
	\label{fig:dataset}
\end{figure}

Guided by this observation, as illustrated in Fig. \ref{fig:dataset}, we construct AeroDS, a comprehensive benchmark for ASI shadow synthesis and removal. AeroDS benchmark comprises two tailored sub-datasets: AeroDS-Syn for fully-supervised training and validation, and AeroDS-Real for out-of-distribution testing. It contains multi-temporal shadowed and shadow-free imagery collected from representative regions across the world. All original samples have a spatial resolution of $0.3\,\mathrm{m}$ and are initially cropped to $256 \times 256$ pixels.

To facilitate accurate pseudo-shadow mask annotation and unified network training, we further normalize the image size to $512 \times 512$ pixels using Real-ESRGAN \cite{9607421}. This upsampling process is used for resolution normalization and mask annotation, while minimizing changes to the original radiometric characteristics of the imagery.

Initially, to build the foundation for AeroDS benchmark, we collected 2,260 real shadowed images and 2,260 unpaired shadow-free images, accompanied by 2,260 manually annotated pseudo-shadow masks. The pseudo-shadow masks are drawn on shadow-free images to indicate the approximate location and extent of the shadows to be synthesized. These data provide the necessary support for modeling the photometric and spatial characteristics of real ASI shadows.

\begin{figure}[t]
	\includegraphics[width=1\linewidth]{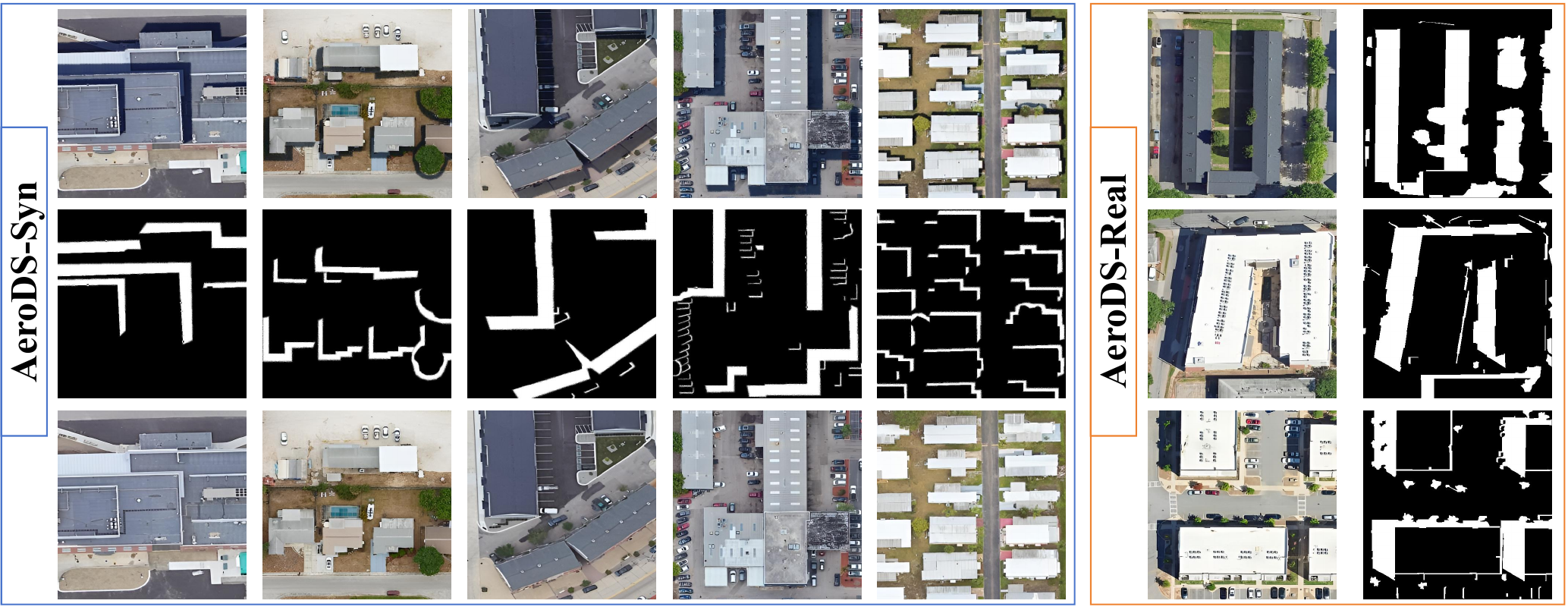}
	\caption{Representative samples from AeroDS-Syn and AeroDS-Real.}
	\label{fig:train_test}
\end{figure}

Based on AeroDS, we construct AeroDS-Syn using the proposed PDSS-Net. Specifically, the shadow-free images and their corresponding pseudo-shadow masks are fed into PDSS-Net, while the collected real shadowed images are used as target-domain references to guide shadow distribution transfer. As shown on the left side of Fig. \ref{fig:train_test}, this process produces paired triplets consisting of a shadow-free image, a synthesized shadow image, and the corresponding shadow mask.

AeroDS-Syn contains 2,260 triplets, among which 2,000 are used for training and the remaining 260 are reserved for testing. Compared with existing ASI shadow datasets, AeroDS-Syn provides high-quality pixel-level supervision with more realistic shadow attenuation and smoother boundary transition characteristics.

To further evaluate the cross-domain generalization capability of deshadowing methods, we additionally construct AeroDS-Real, an independent real-world test set consisting of 260 shadowed ASI samples collected from diverse geographical environments. Since paired shadow-free references are unavailable in real scenes, AeroDS-Real provides manually annotated shadow masks and is used for qualitative evaluation and cross-domain generalization analysis.

Accordingly, AeroDS-Syn supports quantitative evaluation using standard full-reference metrics, whereas AeroDS-Real is employed to assess the visual restoration fidelity and robustness of deshadowing methods in real-world scenarios.

\section{AeroDeshadow Framework} \label{sec:method}

\subsection{PDSS-Net} \label{sec:method1}

\begin{figure*}[t]	
	\includegraphics[width=1\linewidth]{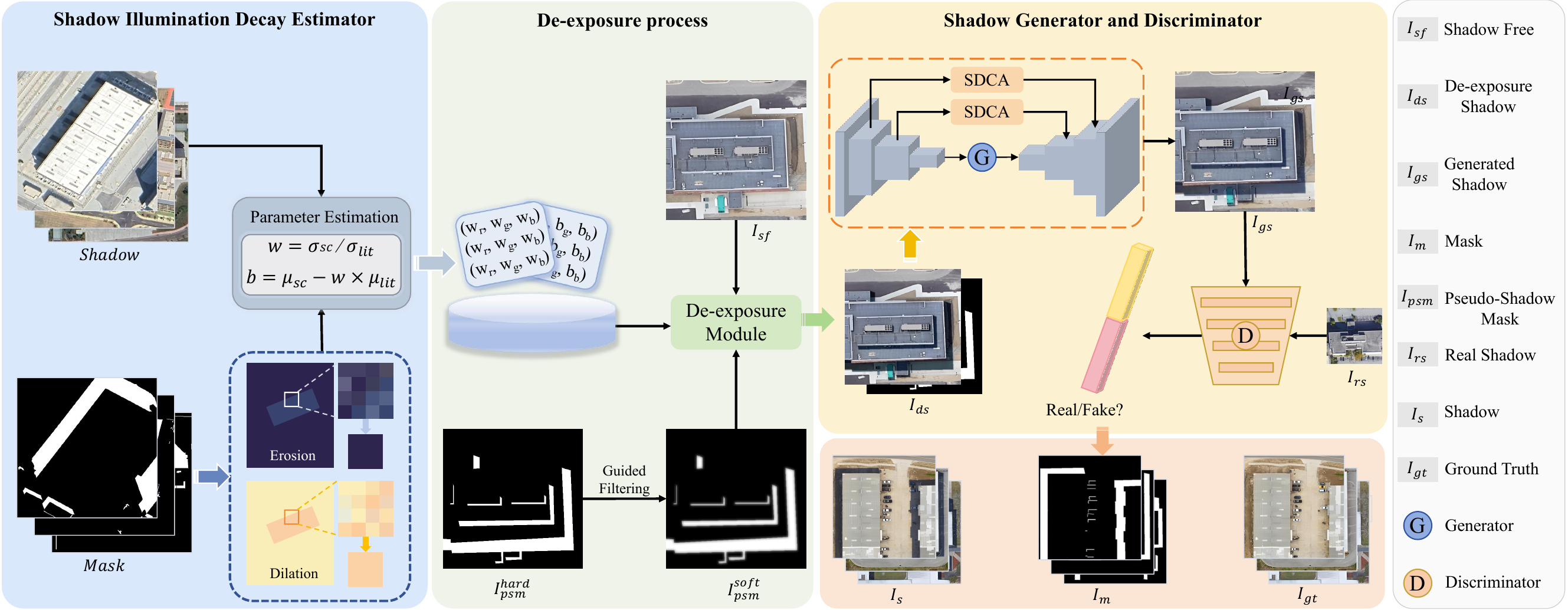}
	\caption{Detailed flowchart of the AeroDS-Syn dataset construction. The process integrates a Shadow Illumination Decay Estimator to extract physical degradation priors $(w, b)$ from real shadows. These parameters drive a physics-guided De-exposure Process for initial synthesis. Finally, a GAN equipped with Spatial-Decay Coordinate Attention (SDCA) refines the initial shadows to simulate realistic spatial attenuation and soft boundaries.} \label{fig:PDSSNet}
\end{figure*}

Distinct from existing methods that treat shadows as homogeneous regions, we formulate shadow modeling as a decoupled physical process where umbra and penumbra follow distinct degradation and restoration dynamics. Building upon this perspective, this section elaborates on the proposed PDSS-Net. The network aims to address the scarcity of paired shadow data in ASI by coupling physical degradation priors with generative learning. As illustrated in Fig. \ref{fig:PDSSNet}, PDSS-Net consists of three core components: (1) a shadow illumination decay estimator, which decouples physically-consistent degradation factors from real ASI; (2) a physics-guided initial synthesis module, which constructs coarse shadow representations based on linear degradation priors; and (3) a refinement generative network integrating the Spatial-Decay Coordinate Attention (SDCA) module to achieve high-fidelity simulation of the physical characteristics of umbra and penumbra regions.

\subsubsection{Shadow Illumination Decay Parameter Extraction}
For a real-world shadowed image $I$, we assume that the shadow pixels $I_{s}$ and their adjacent shadow-free pixels $I_{l}$ follow a linear illumination degradation relationship:
\begin{equation}
	I_{s} = w \cdot I_{l} + b,
\end{equation}
where $w \in \mathbb{R}^3$ represents the channel-wise illumination scaling factor, and $b \in \mathbb{R}^3$ denotes the channel-wise environment light shift bias.

To construct a physically representative illumination decay parameter library, an automated shadow degradation estimator is designed. Given a set of real shadowed images and their corresponding binary masks $I_{m}$, morphological erosion and dilation operations are employed to precisely delineate the shadow core region $I_{sc}$ and the adjacent lit region $I_{lit}$, effectively excluding interference from boundary mixed pixels. Subsequently, the underexposure parameters $(w, b)$ are extracted using statistical moment estimation:
\begin{equation}
	w = \frac{\sigma_{sc}}{\sigma_{lit}}, \quad b = \mu_{sc} - w \cdot \mu_{lit},
\end{equation}
where $\sigma$ and $\mu$ denote the standard deviation and mean vectors of the respective regional features. The resulting parameter library provides authentic physical constraints for subsequent synthesis tasks.

\subsubsection{Physics-Guided Initial Synthesis}
Given a real shadow-free image $I_{free}$ and an annotated pseudo-shadow mask $I_{psm}^{hard}$, a guided filter is employed to smooth the mask since hard masks exhibit prominent step effects at boundaries. This yields an edge-preserving soft mask $I_{psm}^{soft}$:
\begin{equation}
	I_{psm}^{soft} = \text{GuidedFilter}(I_{psm}^{hard}, I_{free}).
\end{equation}

Subsequently, a parameter pair $(w, b)$ is randomly sampled from the prior library to apply a linear attenuation transform to the original image, generating a physics-guided initial shadow image $I_{ds}$:
\begin{equation}
	I_{ds} = I_{free} \odot (\mathbf{1} - I_{psm}^{soft}) + \text{Clip}(w \cdot I_{free} + b) \odot I_{psm}^{soft},
\end{equation}
where $\odot$ denotes element-wise multiplication. This process provides a physically plausible prior-guided initialization input for the network, significantly alleviating the learning difficulty for the generator in complex spectral mappings.

\subsubsection{Shadow Synthesis Network}
As illustrated in Fig. \ref{fig:PDSSNet}, a framework based on CycleGAN \cite{CycleGAN} is adopted to further refine the initial shadow $I_{ds}$ into a realistic shadow image $I_{gs}$, thereby bridging the domain gap between synthetic and real-world data. The generator employs a U-Net \cite{2015unet} style encoder-decoder architecture.

\begin{figure}[t]
	\includegraphics[width=1\linewidth]{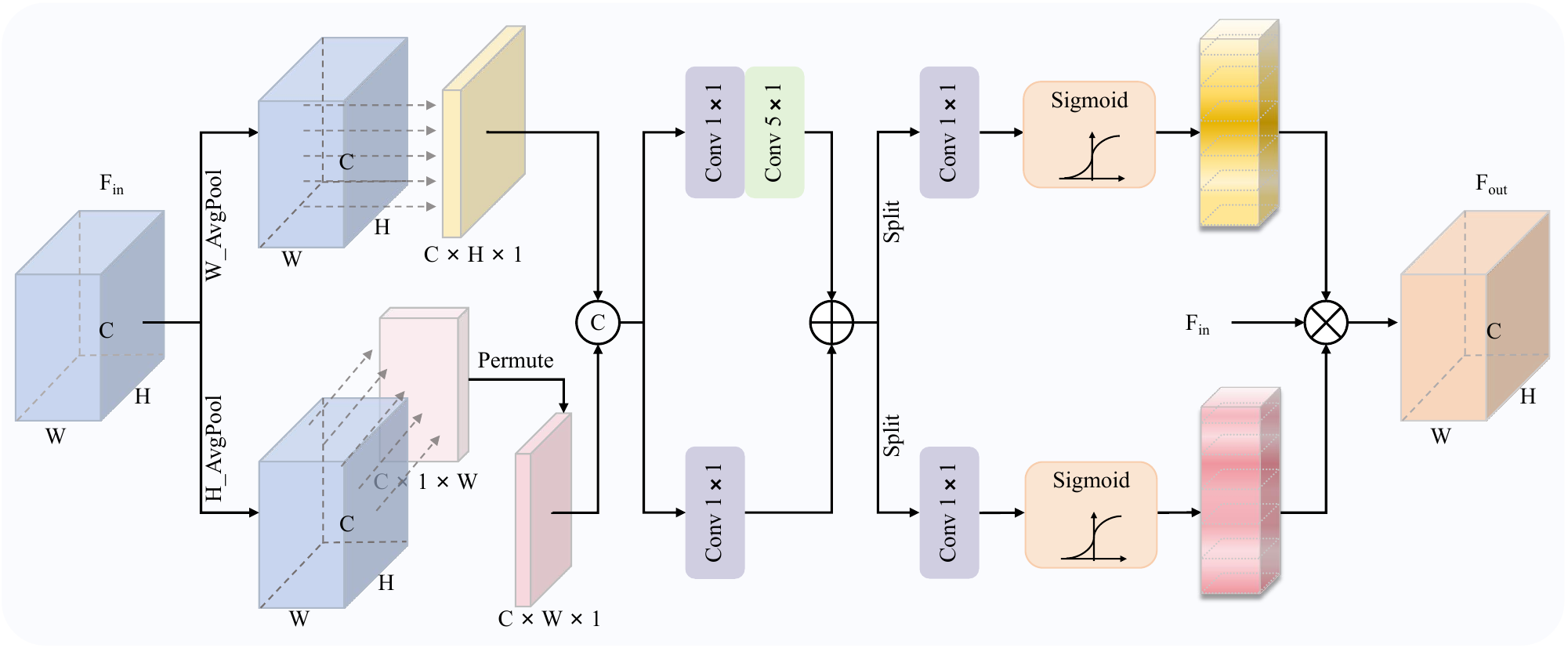}
	\caption{Detailed architecture of the SDCA module. To model the physical transition of shadows, SDCA decouples feature extraction into a local perception branch to preserve umbra edge sharpness, and a decay simulation branch to simulate the non-linear spatial attenuation characteristic of the penumbra.}
	\label{fig:SDCA_Module}
\end{figure}

To enhance the network's capability in modeling the geometric properties of shadows---specifically the smooth transition from umbra to penumbra---we introduce SDCA mechanism at the skip connections. As shown in Fig. \ref{fig:SDCA_Module}, SDCA reweights the input feature $F \in \mathbb{R}^{C \times H \times W}$ via:
\begin{equation}
	F' = F \odot A_h \odot A_w,
\end{equation}
where $A_h, A_w \in [0,1]^{C \times H \times W}$ represent the broadcasted attention maps in the horizontal and vertical directions. 

Specifically, coordinate pooling is performed to preserve spatial positional information:
\begin{equation}
	X_h = \text{AvgPool}_H(F), \quad X_w = \text{AvgPool}_W(F),
\end{equation}
where $X_h \in \mathbb{R}^{N \times C \times H \times 1}$ and $X_w \in \mathbb{R}^{N \times C \times 1 \times W}$. These features are then concatenated to obtain:
\begin{equation}
	Y = \text{Concat}(X_h, X_w^\top) \in \mathbb{R}^{N \times C \times (H+W) \times 1}.
\end{equation}

Subsequently, features are extracted via two parallel branches:
\begin{itemize}
	\item \textit{Local Perception Branch}: A $1 \times 1$ convolution is utilized to maintain the edge sharpness of the umbra region:
	\begin{equation}
		Y_\text{local} = \text{Conv}_{1 \times 1}(Y) \in \mathbb{R}^{N \times M \times (H+W) \times 1},
	\end{equation}
	where $M = \max(8, C / r)$ and $r = 32$ is the channel reduction ratio.
	\item \textit{Decay Simulation Branch}: The feature dimension is first reduced via a $1 \times 1$ convolution, followed by a depthwise $5 \times 1$ strip convolution applied along the spatial dimension to simulate the gradual distance-decay effect of the penumbra transition:
	\begin{equation}
		Y_\text{decay} = \text{StripConv}_\text{1D}(\text{Conv}_{1 \times 1}(Y)).
	\end{equation}
\end{itemize}

The outputs are fused through summation, normalization, and activation:
\begin{equation}
	Y_\text{fused} = \text{BN}(\text{h-swish}(Y_\text{local} + Y_\text{decay})).
\end{equation}

Finally, $Y_\text{fused}$ is split along the spatial dimensions and transformed into the final attention maps $A_h$ and $A_w$ via sigmoid functions:
\begin{equation}
	A_h = \sigma(\text{Conv}_{1 \times 1}(Y_h)), \quad A_w = \sigma(\text{Conv}_{1 \times 1}(Y_w^\top)).
\end{equation}

By balancing local sharpness and decay diffusion, SDCA ensures that the generated shadows exhibit physically consistent characteristics.

\subsubsection{Loss Function Design}
To explicitly align the mathematical formulation with the dataset construction pipeline illustrated in Fig. \ref{fig:PDSSNet}, we denote the initial de-exposure shadow image as $I_{ds}$, the pseudo-shadow mask as $I_{psm}$, and the real shadowed image as $I_{rs}$. The output of our generator $G$ is the refined generated shadow, formulated as $I_{gs} = G(I_{ds}, I_{psm})$. To train PDSS-Net on an unpaired dataset and ensure that the generated shadows possess realistic texture details while strictly preserving the geometric structure of the original scene, the total optimization objective is formulated as:
\begin{equation}
	\mathcal{L}_{PDSS} = \mathcal{L}_{adv} + \lambda_{cyc}\mathcal{L}_{cyc} + \lambda_{back}\mathcal{L}_{back} + \lambda_{idt}\mathcal{L}_{idt},
\end{equation}
where the trade-off parameters are empirically set to $\lambda_{cyc}=10$, $\lambda_{back}=10$, and $\lambda_{idt}=5$. The individual loss components are detailed as follows:

\textbf{Adversarial Loss ($\mathcal{L}_{adv}$)} improves the distributional realism of the generated shadows through a minimax game between the discriminator $D$ and the generator $G$:
\begin{equation}
	\mathcal{L}_{adv} = \mathbb{E}_{I_{rs}}[(D(I_{rs}) - 1)^2] + \mathbb{E}_{I_{ds},I_{psm}}[(D(I_{gs}))^2].
\end{equation}

\textbf{Cycle Consistency Loss ($\mathcal{L}_{cyc}$)} imposes inverse mapping constraints, ensuring content consistency before and after the image transformation:
\begin{equation}
	\begin{split}
		\mathcal{L}_{cyc} &= \mathbb{E}_{I_{ds},I_{psm}}[\|F(I_{gs}) - I_{ds}\|_1] \\
		&\quad + \mathbb{E}_{I_{rs},I_m}[\|G(F(I_{rs}), I_m) - I_{rs}\|_1],
	\end{split}
\end{equation}
where $F$ denotes the inverse generator, $I_m$ represents the real shadow mask corresponding to $I_{rs}$, and $\|\cdot\|_1$ represents the $L_1$ norm.

\textbf{Background Consistency Loss ($\mathcal{L}_{back}$)} utilizes mask constraints to ensure non-shadow region pixels remain invariant:
\begin{equation}
	\mathcal{L}_{back} = \mathbb{E}_{I_{ds},I_{psm}}[\|(I_{ds} - I_{gs}) \odot (\mathbf{1} - I_{psm})\|_1],
\end{equation}
where $\mathbf{1}$ is an all-ones matrix matching the spatial dimensions of the input.

\textbf{Identity Loss ($\mathcal{L}_{idt}$)} constrains the generator to output the original image when a real shadow image is provided as input, thereby preserving the original spectral characteristics:
\begin{equation}
	\mathcal{L}_{idt} = \mathbb{E}_{I_{rs},I_m}[\|G(I_{rs}, I_m) - I_{rs}\|_1].
\end{equation}

\begin{figure*}[t]
	\includegraphics[width=1\linewidth]{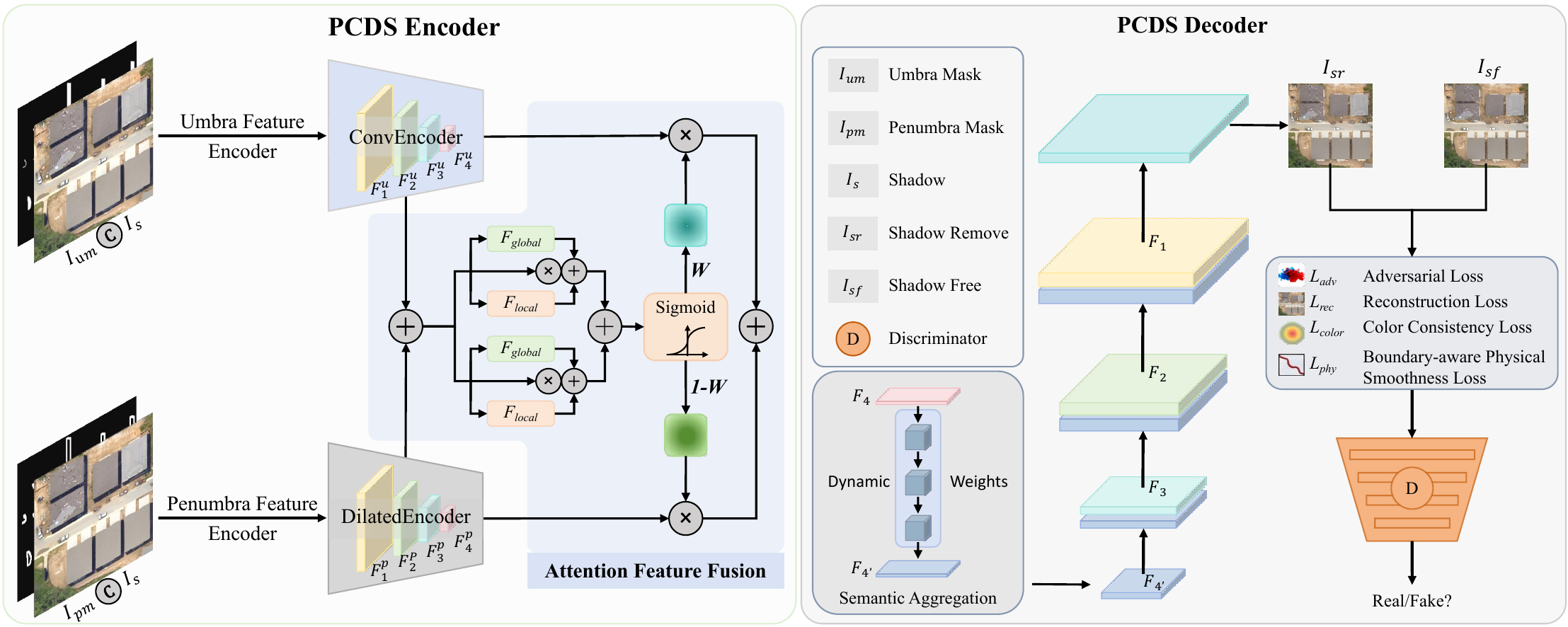}
	\caption{Architecture of the PCDS-Net. The model explicitly decouples the input into umbra and penumbra streams. An Attention Feature Fusion (AFF) module dynamically integrates the high-frequency texture priors from the umbra and the boundary context from the penumbra. A Cascaded Refinement Decoder then progressively reconstructs the shadow-free image from high-level semantics to low-level spatial details.}
	\label{fig:RSGSR}
\end{figure*}

\subsection{PCDS-Net} \label{sec:method2}
This section details the proposed PCDS-Net. Specifically designed to address boundary artifacts and texture loss in ASI shadow removal, PCDS-Net (Fig. \ref{fig:RSGSR}) adopts an end-to-end cascaded architecture comprising three key components: 1) a Penumbra-aware Two-stream Encoder, which decouples the umbra and penumbra branches to respectively anchor the core region texture and the boundary context priors; 2) an Attention Feature Fusion (AFF) module, which utilizes a dynamic weight calibration mechanism to achieve non-linear integration and complementarity of heterogeneous features; 3) a Cascaded Refinement Decoder, which integrates a Semantic Aggregation (SA) module and a multi-scale cascading strategy to guide the reconstruction process toward robust convergence from high-dimensional semantics to low-level spatial details.

\subsubsection{Penumbra-aware Two-stream Encoder}
To achieve differentiated processing,the feature extraction is decoupled into two parallel branches. Using morphological erosion and dilation, an umbra mask $I_{um}$ and a penumbra mask $I_{pm}$ are dynamically derived from the input mask $I_m$.

The \textbf{umbra stream}, termed as the umbra feature encoder, focuses on high-frequency texture reconstruction within the core shadow. It extracts features $F_u^i$ through sequential convolutional blocks:
\begin{equation}
	F_u^i = \begin{cases} 
		\text{ConvBlock}_i([I_s, I_{um}]), & i = 1 \\ 
		\text{ConvBlock}_i(F_u^{i-1}), & i > 1 
	\end{cases},
\end{equation}
where $[ \cdot, \cdot ]$ denotes channel-wise concatenation.

The \textbf{penumbra stream}, defined as the penumbra feature encoder, emphasizes context compensation for boundary transitions. It employs dilated convolutions to expand the receptive field, capturing surrounding background illumination priors. Specifically, it extracts features $F_p^i$ via:
\begin{equation}
	F_p^i = \begin{cases} 
		\text{DilaConvBlock}_i([I_s, I_{pm}]), & i = 1 \\ 
		\text{DilaConvBlock}_i(F_p^{i-1}), & i > 1 
	\end{cases}.
\end{equation}

\subsubsection{Attentional Feature Fusion}
To adaptively integrate $F_u^i$ and $F_p^i$, the AFF module is introduced. For the initial summation $X = F_u^i + F_p^i$, AFF captures local context via point-wise convolutions ($\mathcal{F}_{local}$) and global context via Global Average Pooling ($\mathcal{F}_{global}$). The dynamic weight $W$ is computed as:
\begin{equation}
	W = \sigma\left(\mathcal{F}_{local}(X) + \mathcal{F}_{global}(\text{GAP}(X))\right),
\end{equation}
and the final fused feature is obtained by:
\begin{equation}
	F_{fuse}^i = F_u^i \odot W + F_p^i \odot (\mathbf{1} - W).
\end{equation}
This mechanism ensures the network prioritizes texture restoration in the shadow core while enforcing smooth transitions at boundaries.

\subsubsection{Cascaded Refinement Reconstruction}
In the decoding stage, the deepest feature is first refined by the SA module to generate robust context semantics $F_{sem}$: 
\begin{equation}
	F_{sem} = \mathcal{F}_{DSA}(F_{fuse}^4).
\end{equation}
Subsequently, a cascaded strategy is adopted. At each decoding level $k$, the upsampled feature from the previous stage is concatenated with the fused feature $F_{fuse}$ of the corresponding scale, followed by refinement convolutions:
\begin{equation}
	F_{rec}^k = \mathcal{F}_{conv}\left(\text{Concat}(\text{Up}(F_{rec}^{k-1}), F_{fuse}^{4-k+1})\right),
\end{equation}
where $\text{Up}(\cdot)$ denotes bilinear upsampling. This ensures shallow spatial details continuously calibrate deep semantics, generating sharp and realistic shadow-free images.

\subsubsection{Loss Function Design}
To explicitly align the mathematical formulation with the network architecture illustrated in Fig. \ref{fig:RSGSR}, we denote the input shadowed image as $I_{s}$, the shadow mask as $I_{m}$, and the ground-truth shadow-free image as $I_{sf}$. The output of our generator $G$ is the predicted shadow-free image, formulated as $I_{sr} = G(I_s, I_m)$. To ensure that PCDS-Net accurately reconstructs texture details in the umbra and seamlessly suppresses boundary artifacts in the penumbra, the total optimization objective is formulated as:
\begin{equation}
	\mathcal{L}_{PCDS} = \mathcal{L}_{adv} + \mathcal{L}_{rec} + \lambda_c \mathcal{L}_{color} + \lambda_p \mathcal{L}_{phy},
\end{equation}
where the trade-off hyper-parameters $\lambda_{c}$ and $\lambda_{p}$ are empirically set to $200$ and $10$, respectively. The individual losses are detailed as follows:

\textbf{Adversarial Loss ($\mathcal{L}_{adv}$)} improves the distributional realism of the restored results via a minimax game:
\begin{equation}
	\mathcal{L}_{adv} = \mathbb{E}_{I_{sf}}[(D(I_{sf}) - 1)^2] + \mathbb{E}_{I_{s},I_{m}}[(D(I_{sr}))^2].
\end{equation}

\textbf{Reconstruction Loss ($\mathcal{L}_{rec}$)} ensures consistency with the ground truth in terms of both pixel values and high-level semantic features:
\begin{equation}
	\mathcal{L}_{rec} = \lambda_{L1}\|I_{sr} - I_{sf}\|_1 + \lambda_{per} \sum_{k} \|\phi_k(I_{sr}) - \phi_k(I_{sf})\|_1,
\end{equation}
where $\phi_k$ denotes the feature map of the $k$-th layer of a pre-trained VGG-19 network, with weighting coefficients $\lambda_{L1}=80$ and $\lambda_{per}=7$.

\textbf{Color Consistency Loss ($\mathcal{L}_{color}$)} preserves the inter-channel intensity ratios to maintain spectral fidelity. For each pixel, the channel-wise ratio vector is computed by normalizing the RGB channels against their sum, and the loss is defined as the mean absolute error between the predicted and ground-truth ratio maps:
\begin{equation}
	\mathcal{L}_{color} = \frac{1}{N}\sum_{p} \left\| \frac{I_{sr, p}}{\sum_c I_{sr, p}^c + \epsilon} - \frac{I_{sf, p}}{\sum_c I_{sf, p}^c + \epsilon} \right\|_1,
\end{equation}
where $c \in \{R, G, B\}$ denotes the color channel index, $p$ is the pixel index, $N$ is the total number of pixels, and $\epsilon$ is a small constant for numerical stability.

\textbf{Boundary-aware Physical Smoothness Loss ($\mathcal{L}_{phy}$)} explicitly focuses the generative capacity on eliminating visual discontinuities at shadow boundaries. Driven by the physical Retinex model, we approximate the illumination map as $L_{est} = I_{s} / (I_{sr} + \epsilon)$, where $\epsilon$ is a small constant for numerical stability. The loss is computed as:
\begin{equation}
	\mathcal{L}_{phy} = \sum \|\nabla L_{est} \odot (\mathbf{1} - I_{pm})\|_1 + \sum \|\nabla I_{sr} \odot I_{pm}\|_1,
\end{equation}
where $\nabla$ represents the gradient operator. This ensures spatial smoothness of the illumination in non-penumbra regions and texture consistency within the penumbra transition zone.

\section{Experiments and Analysis}
This section details the implementation specifics and the experimental environment of the proposed framework. Through large-scale comparative experiments on multiple benchmark datasets, we systematically evaluate the shadow generation performance of PDSS-Net and the superiority of PCDS-Net in shadow removal. Furthermore, quantitative ablation studies are conducted to verify the individual contributions of each core component to the overall model performance.

\subsection{Experimental Settings}

\subsubsection{Implementation Details}
Both PDSS-Net and PCDS-Net were implemented using the PyTorch framework and trained on an NVIDIA RTX 3090 Ti GPU. The optimization process employed the Adam optimizer with an initial learning rate of $2 \times 10^{-4}$, complemented by a linear decay scheduling strategy. The batch sizes for PDSS-Net and PCDS-Net were set to 2 and 4, with training cycles of 100 and 200 epochs, respectively. To ensure rich spatial texture information and effective extraction of cross-scale features, all input images were uniformly resized or cropped to $512 \times 512$ pixels. 

\subsubsection{Datasets}
We evaluate the proposed framework on three representative ASI datasets:\\
\textbf{AeroDS}: As the core contribution, AeroDS introduces the AeroDS-Syn subset, which contains 2,000 high-quality paired triplets for training. Additionally, it establishes a dual-track testing set comprising 260 synthetic triplets (AeroDS-Syn Test) and 260 real-world samples (AeroDS-Real Test). This composition enables comprehensive validation across both controlled settings and complex real-world environments.\\
\textbf{AISD \cite{LUO2020443}}: Although originally designed for shadow detection, this dataset was repurposed for shadow removal in our study. We conducted super-resolution reconstruction on 514 original shadowed images to construct a paired training set, and reserved 51 test samples to evaluate cross-scene generalization.\\
\textbf{SRGTA \cite{chu2024gta}}: A large-scale virtual synthetic dataset containing 1,000 ``shadow/mask/shadow-free'' triplets. Following the official protocol, 930 samples are used for training and 70 samples for testing.

\subsubsection{Competitors}
To ensure the rigorousness of the experimental conclusions, we conduct comprehensive comparisons for both synthesis and removal tasks:\\
\textbf{Shadow Synthesis}: Five representative methods spanning different paradigms are selected: SynShadow \cite{inoue2021learning}, Mask-ShadowGAN \cite{hu2019mask}, G2R-ShadowNet \cite{G2R_ShadowNet}, UP-ShadowGAN \cite{UPShadowGAN}, and RS-GSSR \cite{10967107}.\\
\textbf{Shadow Removal}: Eight SOTA algorithms are selected, including traditional physics-driven methods (RSISR \cite{SILVA2018104}, MAOSD \cite{ZHANG2025127769}) and deep learning architectures covering CNNs, GANs, and Transformers (Mask-ShadowGAN \cite{hu2019mask}, G2R-ShadowNet \cite{G2R_ShadowNet}, Self-ShadowGAN \cite{Jiang2023}, ESCNet \cite{luo2022uavsc}, UP-ShadowGAN \cite{UPShadowGAN}, and RS-GSSR \cite{10967107}). This selection ensures a comprehensive benchmarking across both traditional physical-model and modern data-driven architectures. 

All competing methods were retrained using their officially released source codes under optimal parameter configurations.

\subsubsection{Evaluation Metrics}
For shadow synthesis, we introduce two physically interpretable metrics: the Shadow-to-Lit Intensity Ratio (SLR) and the chromaticity deviation in the $a$ channel ($\Delta a$). The SLR is defined as the ratio of the mean luminance between the shadowed region and its adjacent non-shadowed region. Comparing the statistical distributions (SLR range) of synthetic and real samples enables a quantitative assessment of whether the generative model accurately captures the illumination attenuation laws inherent in specific ASI. The $\Delta a$ measures the variation of the $a$ component in the CIELab color space before and after shadow synthesis. Fundamentally, shadow formation stems from the reduction of incident light intensity rather than an alteration of the intrinsic surface reflectance. Since the $a$ channel represents the red-green opponent color information and exhibits high robustness to variations in light intensity, a physically consistent shadow synthesis process should maintain the stability of the $a$.

For shadow removal, in synthetic scenarios with available ground truth (GT), Peak Signal-to-Noise Ratio (PSNR), Structural Similarity (SSIM) \cite{wang2004ssim}, and Root Mean Square Error (RMSE) are adopted to measure reconstruction quality. In real-world ASI scenarios without GT, we introduced a multi-dimensional evaluation system, including Image Entropy and Blind/Referenceless Image Spatial Quality Evaluator (BRISQUE). This comprehensive evaluation scheme ensures that the algorithm is objectively assessed in terms of both physical accuracy and perceived quality.

\subsection{Experimental Results}

\begin{table}[t] \centering
\caption{Quantitative Evaluation of Shadow Synthesis Methods on the AeroDS-Syn Test Set. Proximity to the ``Real reference'' distribution indicates superior physical fidelity. Best \\ results are \textbf{bold}; second-best are \underline{underlined}.} \label{table_shadow_synthesis}
\renewcommand{\arraystretch}{1.2}
\resizebox{\columnwidth}{!}{
\begin{tabular}{lcccc}	\toprule
Method & Publication & MeanSLR & SLR Range & $\Delta a$ \\ 	\midrule
Real Reference & --- & 0.366 & (0.282, 0.436) & -0.385 \\ \midrule
SynShadow \cite{inoue2021learning}& TCSVT 2021 & 0.293 & (0.104, 0.513) & -1.560 \\ 
Mask-ShadowGAN \cite{hu2019mask} & ICCV 2019 & 0.719 & (0.506, 0.967) & \underline{-0.619} \\ 
G2R-ShadowNet \cite{G2R_ShadowNet} & CVPR 2021 & 0.142 & (0.051, 0.259) & 0.315 \\ 
UP-ShadowGAN \cite{UPShadowGAN} & TGRS 2024 & 0.668 & (0.377, 1.064) & \textbf{-0.177} \\ 
RS-GSSR \cite{RS_GSSR} & TGRS 2025 & \underline{0.346} & \underline{(0.224, 0.491)} & 0.959 \\ 
\rowcolor[gray]{0.9} \textbf{PDSS-Net} & \textbf{Ours} & \textbf{0.353} & \textbf{(0.219, 0.519)} & 0.803 \\  \hline
\end{tabular}}
\end{table}

\subsubsection{Experimental Results on Shadow Synthesis}
Table \ref{table_shadow_synthesis} summarizes the quantitative comparison between the proposed PDSS-Net and several SOTA shadow synthesis algorithms on the AeroDS-Syn test set. The experimental results demonstrate that PDSS-Net yields a meanSLR and SLR Range that align with the real reference, indicating its synthesized illumination attenuation accurately reflects real-world distributions. Physical analysis reveals that deviations in meanSLR directly impact visual realism. For instance, the significantly low meanSLR of G2R-ShadowNet leads to an excessive intensity degradation in shadowed regions, which obscures the underlying ground textures. Conversely, the high meanSLR of Mask-ShadowGAN results in a visual bias resembling semi-transparent artifacts, lacking the physical density of actual occlusions. PDSS-Net exhibits a broader distribution in the SLR Range metric, providing strong evidence that the stochastic de-exposure module accurately simulates the luminance diversity caused by atmospheric conditions and varying illumination angles. 

While UP-ShadowGAN and Mask-ShadowGAN achieve $\Delta a$ values closer to the real reference, this proximity originates from their global transfer mechanisms. By applying uniform chromatic shifts, these methods optimize global averages at the expense of introducing spatial mismatches and color bleeding. Inherently, $\Delta a$ evaluates global chromaticity drift, making it insensitive to fine-grained local variations. To preserve structural boundaries, PDSS-Net explicitly models the non-linear spatial attenuation of the penumbra via the SDCA mechanism. The resulting localized gradients inherently introduce numerical variance when aggregated into a global metric like $\Delta a$. As consistently evidenced by the optimal meanSLR, PDSS-Net prioritizes physically accurate local illumination decay and robust geometric fidelity over uniform global chromatic shifts.

\begin{figure*}[t]
	\includegraphics[width=1\linewidth]{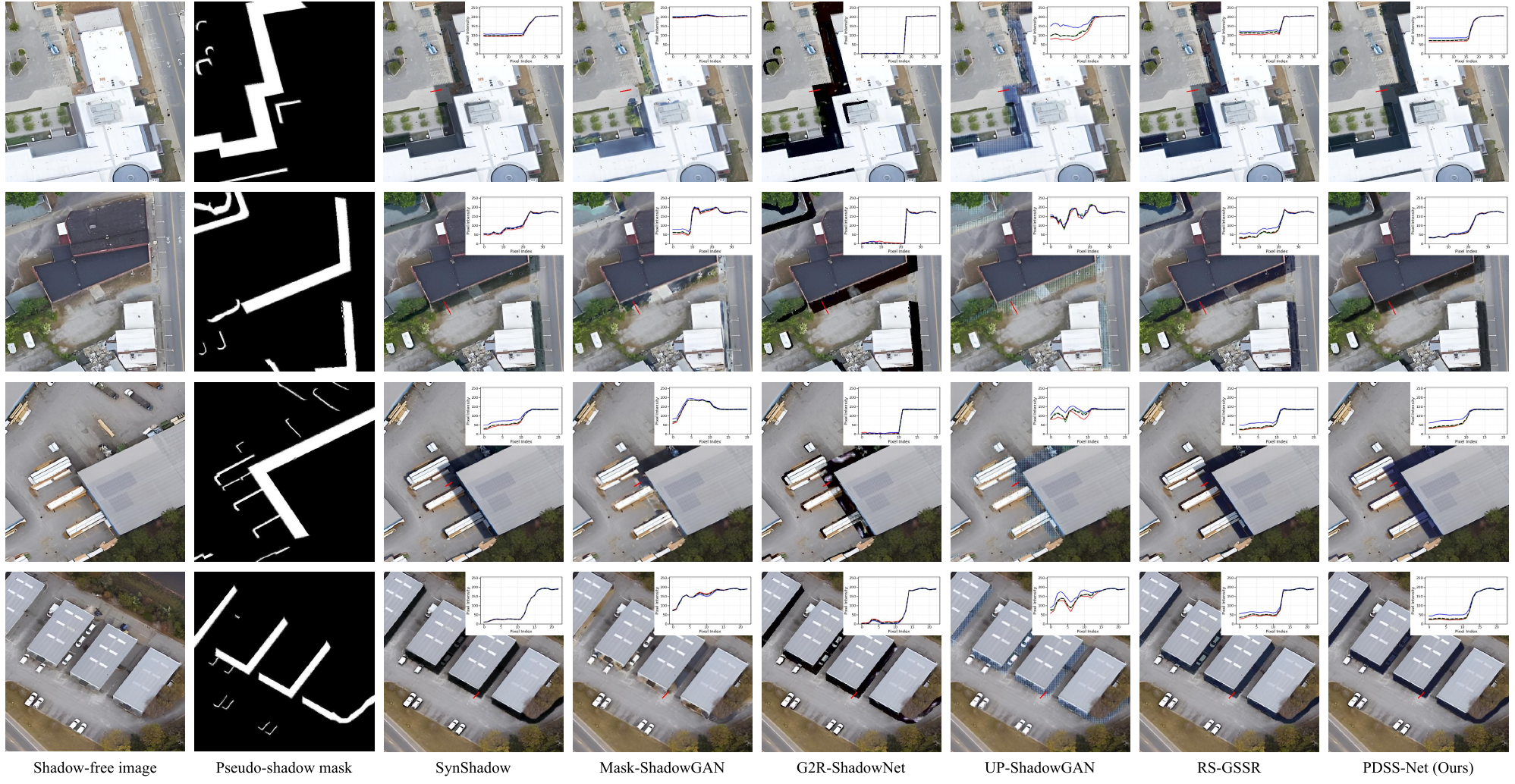}
	\caption{Qualitative comparison of shadow synthesis methods, accompanied by pixel intensity profile analysis across shadow boundaries. The profile curves illustrate the transition from the shadowed region to the background. Compared to the step-like transitions or intensity fluctuations observed in other methods, PDSS-Net (h) produces a smooth, non-linear attenuation highly consistent with physical optical scattering.}
	\label{fig:SG_Compare}
\end{figure*}

Fig. \ref{fig:SG_Compare} provides a visual comparison of the synthesis results in ASI scenes along with the corresponding profile analysis of boundary pixels. According to the principles of physical optics, the transition region of a real shadow should exhibit a non-linear and smooth evolution. Linear methods, such as SynShadow, rely on simplistic linear decay models; despite the introduction of post-filtering, its fixed degradation factors are inadequate to characterize the complex radiative transfer environment in ASI. Due to the lack of explicit spatial geometric constraints, Mask-ShadowGAN and UP-ShadowGAN primarily rely on style transfer learning, which often causes significant chromatic aberration in the red and green channels, resulting in unnatural blue-greenish shadow tones. 

On the other hand, methods like G2R-ShadowNet and RS-GSSR rely on local cropping to construct unpaired training samples, a strategy that limits the preservation of global texture continuity and often yields synthesized shadows with uniform intensity distributions. Although RS-GSSR incorporates blurring to soften boundary gradients, its intensity profiles still indicate step-like transitions, which differ from the gradual umbra-to-penumbra evolution observed in natural settings. In contrast, the proposed PDSS-Net processes the full spatial context of the ASI. By integrating the SDCA module, the model captures spatial decay patterns during feature extraction, facilitating a more physically grounded simulation of spatial attenuation. As reflected in the profile analysis, the transition regions generated by PDSS-Net exhibit non-linear attenuation curves that align closely with the photometric properties of real shadows, demonstrating the utility of the proposed architecture in physical shadow modeling.

\subsubsection{Experimental Results on Shadow Removal}
\begin{table*}[ht]	\centering
\caption{Quantitative Comparison of Shadow Removal Performance Across AeroDS, AISD, and SRGTA Datasets. \\ Full-reference metrics (PSNR, SSIM, RMSE) are evaluated on synthetic data, while no-reference \\ metrics (Entropy, BRISQUE)  are used for real-world generalization assessment.}
\label{table_shadow_removal_results}
\renewcommand{\arraystretch}{1.2} 
\resizebox{\textwidth}{!}{%
\begin{tabular}{cc ccc cc cc ccc} \toprule
\multirow{3}{*}{\textbf{Method}} & \multirow{3}{*}{\textbf{Publication}}   & \multicolumn{3}{c}{AeroDS-Syn Test} & \multicolumn{2}{c}{AeroDS-Real} & \multicolumn{2}{c}{AISD Test} & \multicolumn{3}{c}{SRGTA Test} \\ 
\cmidrule(lr){3-5} \cmidrule(lr){6-7} \cmidrule(lr){8-9} \cmidrule(lr){10-12}
&  & PSNR-S $\uparrow$ & SSIM-S $\uparrow$ & RMSE-S $\downarrow$ & Entropy $\uparrow$ & BRISQUE $\downarrow$ & Entropy $\uparrow$ & BRISQUE $\downarrow$ & PSNR-S $\uparrow$ & SSIM-S $\uparrow$ & RMSE-S $\downarrow$ \\ \hline

RSISR \cite{SILVA2018104} & ISPRS 2018 & 14.40 & 0.66 & 23.50 & 6.28 & 24.86 & 6.24 & 22.57 & 14.53 & 0.58 & 22.45 \\
Mask-ShadowGAN \cite{hu2019mask} & ICCV 2019 & 16.13 & 0.68 & 18.79 & 6.89 & 32.39 & 6.24 & 27.28 & 14.04 & 0.50 & 23.77 \\
G2R-ShadowNet \cite{G2R_ShadowNet} & CVPR 2021 & 16.05 & 0.55 & 19.85 & 7.10 & 20.24 & 6.69 & 16.27 & 14.93 & 0.55 & 21.17 \\
Self-ShadowGAN \cite{Jiang2023} & IJCV 2023 & 16.48 & 0.70 & 18.70 & \underline{7.33} & 18.93 & \underline{7.07} & 19.22 & 15.54 & 0.59 & 20.92 \\
ESCNet \cite{luo2022uavsc} & TGRS 2023 & 15.06 & 0.35 & 20.54 & \underline{7.33} & 24.01 & 6.96 & 24.44 & 20.26 & 0.74 & 12.60 \\
UP-ShadowGAN \cite{UPShadowGAN} & TGRS 2024 & 11.98 & 0.46 & 28.52 & 6.94 & 31.86 & 6.01 & 24.88 & 15.42 & 0.55 & 20.49 \\
MAOSD \cite{ZHANG2025127769} & ESWA 2025 & 15.98 & 0.66 & 18.66 & \underline{7.33} & 24.01 & \underline{7.07} & 18.91 & 15.15 & 0.59 & 21.18 \\
RS-GSSR \cite{10967107} & TGRS 2025 & \underline{20.25} & \underline{0.73} & \underline{12.88} & 7.32 & \textbf{11.84} & 6.94 & \underline{13.80} & \underline{21.57} & \textbf{0.80} & \underline{11.28} \\ 
\rowcolor[gray]{0.9} \textbf{PCDS-Net(Ours)} & \textbf{-} & \textbf{21.91} & \textbf{0.79} & \textbf{11.35} & \textbf{7.40} & \underline{12.70} & \textbf{7.17} & \textbf{11.56} & \textbf{22.37} & \underline{0.78} & \textbf{9.41} \\ \hline
\end{tabular}
}
\end{table*}

Table \ref{table_shadow_removal_results} summarizes the quantitative evaluation results of the proposed PCDS-Net compared with eight SOTA shadow removal methods across four representative datasets. The experimental results demonstrate that on the AeroDS-Syn test set, PCDS-Net achieves the optimal performance across three key metrics: PSNR, SSIM \cite{wang2004ssim}, and RMSE within the shadowed regions, proving its superior efficacy in pixel-level reconstruction accuracy and structural consistency. For the SRGTA high-resolution virtual scenes, considering the variations in GPU memory load across different models, we uniformly adopt an overlapping sliding window inference strategy for image stitching to minimize evaluation bias. Under these stringent conditions, our method still maintains the best or second-best performance across all metrics. 

To further verify the cross-domain transferability, we employ an out-of-distribution generalization scheme by training on synthetic data and testing it on real-world datasets. On the AeroDS-Real test set and the AISD dataset, PCDS-Net exhibits significant advantages in Entropy and BRISQUE. This indicates that while recovering latent texture information in shadowed areas, our method effectively suppresses artifact diffusion and improves the perceptual quality and spectral fidelity of real-world ASI. In summary, PCDS-Net precisely fits the light-shadow transformations in synthetic scenarios, and possesses robust cross-domain parameter transferability.

\begin{figure*}[t] \centering
	\includegraphics[width=1\linewidth]{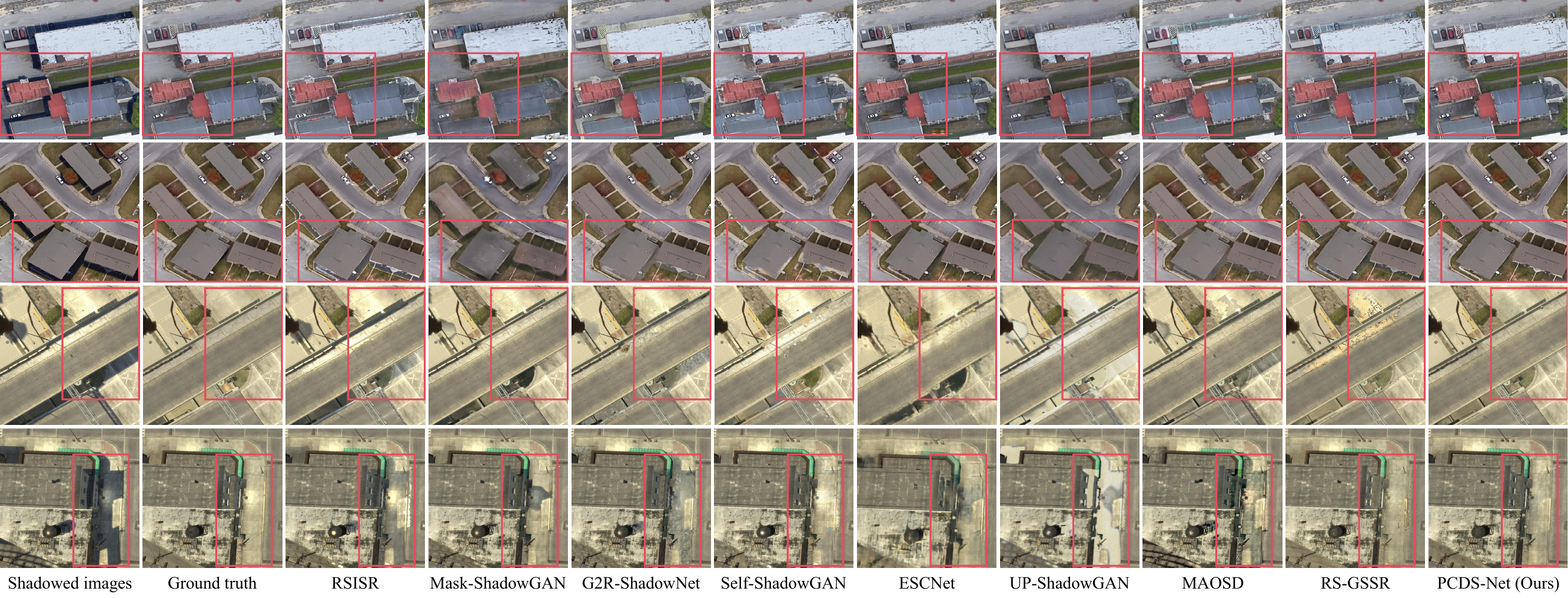}
	\caption{Qualitative comparison of shadow removal performance on the AeroDS-Syn (top two rows) and SRGTA (bottom two rows) datasets. Boxes indicate challenging boundary regions and complex ground textures. PCDS-Net (k) mitigates the spatial mismatches and local over-exposure as observed in several competing methods, yielding visually continuous transitions and maintaining robust geometric fidelity.}	\label{fig:AeroDS_Syn}
\end{figure*}

Fig. \ref{fig:AeroDS_Syn} presents the qualitative comparison results on synthetic and simulated datasets. As observed in Fig. \ref{fig:AeroDS_Syn}(c) and (i), traditional methods, due to their heavy reliance on simplistic linear gain models, succeed in increasing the brightness of shadowed areas but are prone to inducing local over-exposure. This leads to obvious visual discontinuities between the corrected regions and the surrounding background. In contrast, existing deep learning methods generally face issues such as spectral shifts, residual artifacts, and over-sharpening. 

Specifically, Mask-ShadowGAN and UP-ShadowGAN often result in erroneous compensation of non-shadowed areas and severe residual artifacts due to the lack of mask geometric guidance. G2R-ShadowNet and RS-GSSR lose global spatial context by employing patch-based training strategies, leading to blurred detail representation in the restored regions. Furthermore, Self-ShadowGAN and ESCNet exhibit limitations in handling the non-uniform transition effects in the penumbra. Benefiting from the model advantages of the umbra-penumbra decoupling design, PCDS-Net achieves seamless transitions at shadow boundaries while recovering high-frequency details, maintaining stable illumination consistency.

\begin{figure*}[t]	\centering
	\includegraphics[width=1\linewidth]{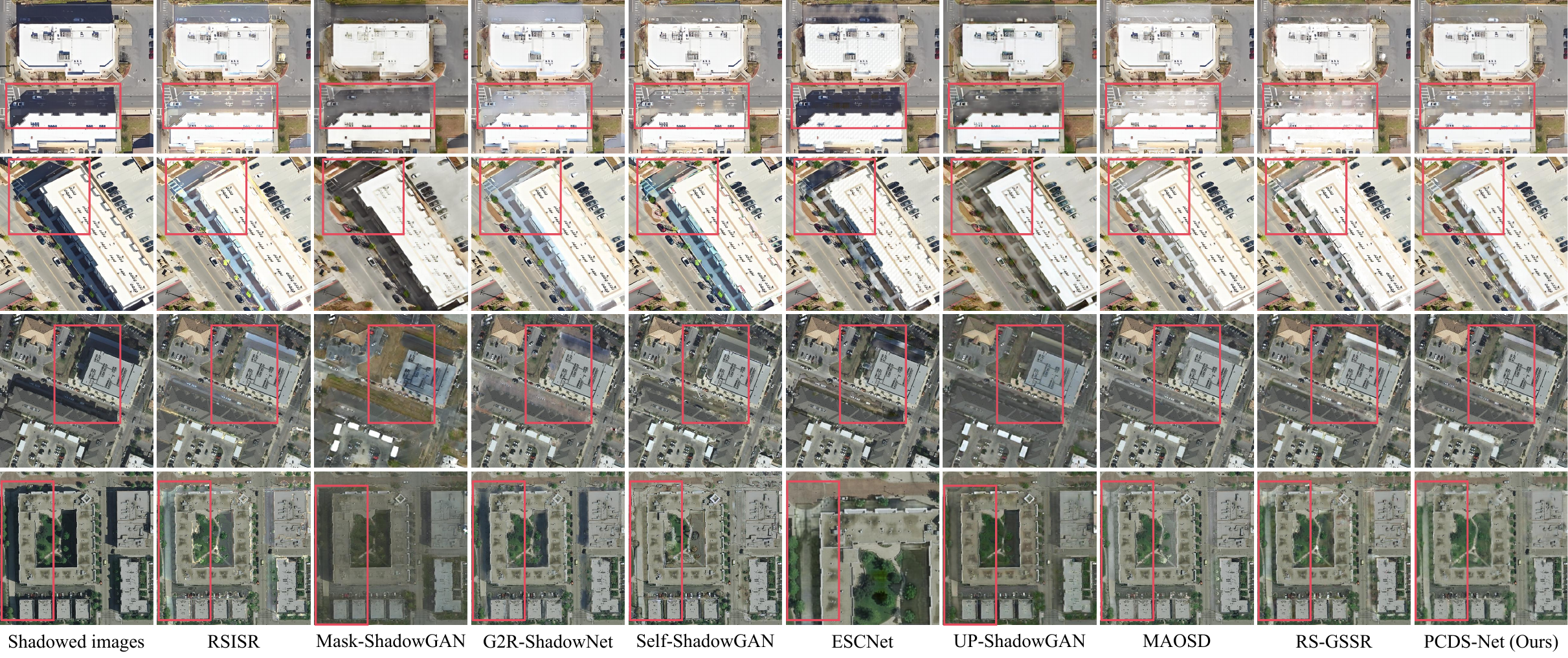}
	\caption{Qualitative comparison of shadow removal performance on real-world ASI from the AeroDS-Real (top two rows) and AISD (bottom two rows) datasets. To evaluate cross-domain generalization, all models were trained on synthetic data. Red boxes highlight complex scenarios where shadows interweave with vegetation or irregular street structures. While competing methods frequently suffer from blue-greenish color distortions or boundary fractures in these uncontrolled environments, the proposed PCDS-Net robustly suppresses environmental color interference and recovers physically consistent transitions.} \label{fig:SRC_test_real-test51}
\end{figure*}

To evaluate the generalization performance in uncontrolled environments, the model trained solely on synthetic datasets is directly applied to real-world ASI, with the comparative results shown in Fig. \ref{fig:SRC_test_real-test51}. As seen in the first and fourth rows of Fig. \ref{fig:SRC_test_real-test51}, competing methods exhibit varying degrees of chromatic aberrations or under-compensation in real-world scenes. 

In contrast, the results processed by PCDS-Net are more consistent with natural illumination distributions. As shown in the second row of Fig. \ref{fig:SRC_test_real-test51}, at complex edges where vegetation and shadows interweave, competing methods are easily disturbed by green vegetation features, resulting in severe blue-greenish color distortions in the restored light. Some methods suppress color bias but introduce significant step-like artifacts in the penumbra. Leveraging its penumbra-aware branch, PCDS-Net achieves a physically consistent and smooth transition from the umbra to the background while suppressing environmental color interference. In the complex street scenes in the third row, while competing methods suffer from both blurred details and boundary fractures, PCDS-Net maintains clear surface textures. Overall, PCDS-Neteffectively addresses the transferability challenges of deep learning models, highlighting its robust applicability to real-world ASI.

\subsection{Ablation Studies}

\subsubsection{Ablation Study of PDSS-Net}

\begin{table}[t] \centering
\caption{Ablation Study of PDSS-Net on AeroDS-Syn Dataset}
\label{table_ablation1}
\renewcommand{\arraystretch}{1.2}
\begin{tabular}{lccc} \toprule
Method & MeanSLR & SLR Range & $\Delta a$ \\  \midrule
Real reference & 0.366 & (0.282, 0.436) & -0.385 \\  
w/o DEP & 0.296 & (0.220, 0.369) & \textbf{0.105} \\ 
\rowcolor[gray]{0.9} \textbf{PDSS-Net} & \textbf{0.353} & \textbf{(0.219, 0.519)} & 0.803 \\ \hline
\end{tabular}
\end{table}
	
\begin{figure}[t] \centering
\includegraphics[width=1\linewidth]{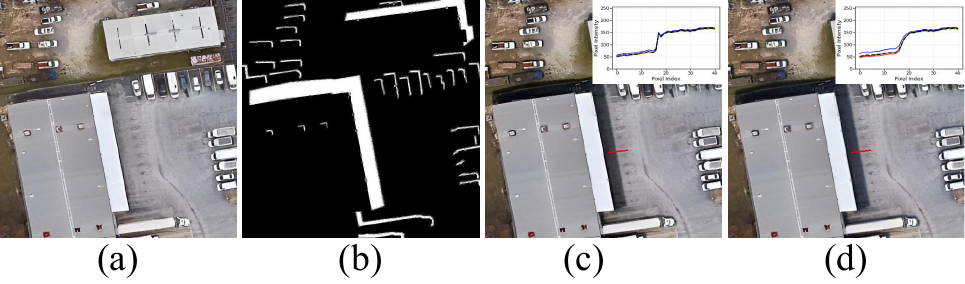}
\caption{Visual ablation analysis of the SDCA module within PDSS-Net. The removal of SDCA (c) leads to sharp boundary mutations and intensity ringing. The full framework (d) is observed to recover the gradual spatial decay of the penumbra. (a) Shadow-free image. (b) Pseudo-shadow mask. (c) Results without the SDCA module. (d) Full PDSS-Net framework.}
\label{fig:Ablation_SDCA}
\end{figure}

To evaluate the contributions of key components within PDSS-Net to the physical realism of synthesized shadows, we conducted an ablation study while maintaining a constant encoder-decoder backbone.

First, we investigated the pivotal role of the De-exposure Processing (DEP) module in simulating realistic light-shadow mapping. The quantitative results in Table \ref{table_ablation1} indicate that without the DEP module, although the model exhibits a lower $\Delta a$ value, its meanSLR significantly deviates from the real reference, and the SLR Range shrinks considerably. This numerical trend suggests that a learning process lacking physical prior guidance struggles to capture the complex illumination attenuation laws inherent in ASI. In contrast, the complete PDSS-Net with DEP facilitates a physics-driven warm-start initialization. The generated shadows achieve higher consistency with real scenes in terms of luminance dynamic range, thereby enhancing the physical diversity of the synthetic data.

Subsequently, we explored the influence of the SDCA module on the transition characteristics of the penumbra. As illustrated in Fig. \ref{fig:Ablation_SDCA}, the removal of SDCA leads to severe intensity oscillations and visual discontinuities at shadow boundaries. Upon integrating the SDCA mechanism, the model precisely captures the spatial degradation laws diffusing from the umbra core to the surrounding non-shadowed regions. 
This ensures that the synthesized shadows exhibit a smooth, non-linear attenuation trend consistent with physical optical scattering in the transition zones.

\subsubsection{Ablation Study of PCDS-Net}
\begin{table}[t] \centering \footnotesize 
\setlength{\tabcolsep}{3pt} 
\caption{Ablation Study of PCDS-Net on AeroDS Dataset}
\label{table_ablation2}
\begin{tabular}{cccccccc}	\toprule
\multirow{3}{*}{UFE} & \multirow{3}{*}{PFE} & \multirow{3}{*}{AFF} & \multicolumn{3}{c}{AeroDS-Syn} & \multicolumn{2}{c}{AeroDS-Real} \\
\cmidrule(lr){4-6} \cmidrule(lr){7-8}
& & & PSNR-S$\uparrow$ & SSIM-S$\uparrow$ & RMSE-S$\downarrow$ & Entropy$\uparrow$ & BRISQUE$\downarrow$ \\  \midrule
\checkmark &   &   & 19.69 & 0.73 & 13.84 & \underline{7.45} & 25.02 \\
& \checkmark &     & 18.41 & 0.66 & 16.18 & \textbf{7.53} & \underline{22.94} \\
\checkmark & \checkmark &            
& \underline{20.82} & \underline{0.78} & \underline{13.06} & 7.29 & 23.98 \\
\rowcolor[gray]{0.9} \checkmark & \checkmark & \checkmark 
& \textbf{20.91} & \textbf{0.79} & \textbf{11.35} & 7.40 & \textbf{12.70} \\ \hline
\end{tabular}
\end{table}

\begin{figure}[t]
	\centering
	\includegraphics[width=1\linewidth]{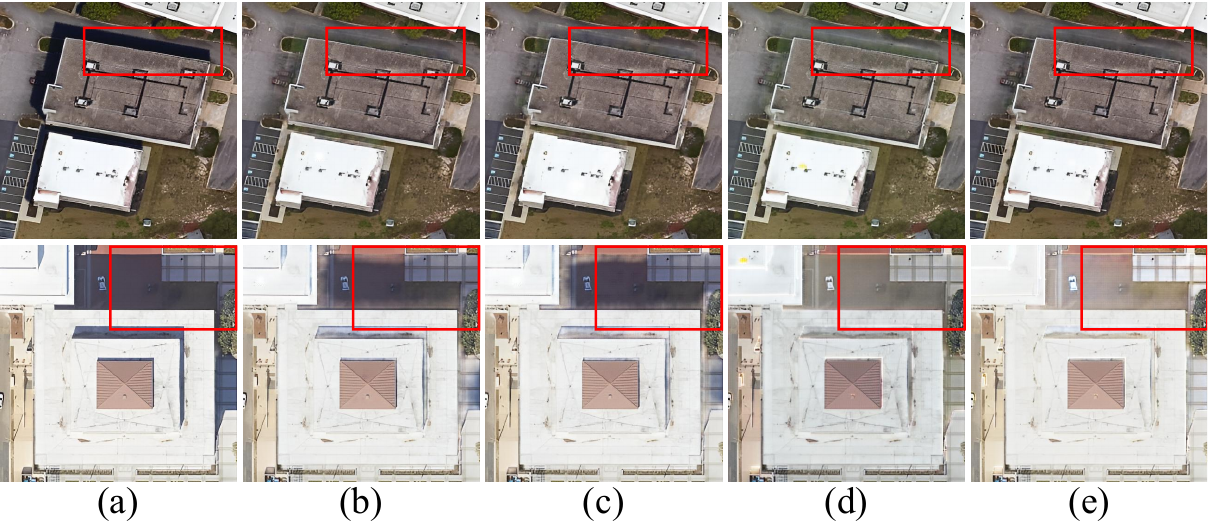}
	\caption{Qualitative visual comparison of the ablation study for the proposed PCDS-Net. (a) Shadow images. (b) Result with UFE only. (c) Result with PFE only. (d) Result without the AFF module (replaced by simple concatenation). (e) Full PCDS-Net framework.} \label{fig:Ablation_SR}
\end{figure}

To validate the individual contributions of PCDS-Net components to shadow removal, we evaluated the framework on AeroDS-Syn and AeroDS-Real. The experimental results are presented from both quantitative and qualitative perspectives in Table \ref{table_ablation2} and Fig. \ref{fig:Ablation_SR}.

First, we investigated the complementarity between the Umbra Feature Encoder (UFE) and the Penumbra Feature Encoder (PFE). When only UFE is retained, the model achieves partial recovery of the umbra region; however, the BRISQUE metric on real images deteriorates significantly to $25.02$. Physically, the absence of penumbra features limits the receptive field, hindering the reconstruction of gradual illumination transitions at boundaries. Visually, as shown in Fig. \ref{fig:Ablation_SR}(b), the reconstructed images exhibit conspicuous halo artifacts at the original shadow boundaries.

Conversely, when only PFE is utilized, the PSNR-S drops from $21.91$~dB to $18.41$~dB, representing a substantial degradation of $3.50$~dB. 
This quantitative decrease demonstrates that the PFE's large receptive field alone is insufficient to recover intrinsic surface reflectance without the high-frequency texture priors provided by UFE. The UFE is thus verified as indispensable for maintaining structural accuracy in the shadow core, as illustrated in Fig. \ref{fig:Ablation_SR}(c).

Finally, we verify the critical role of the AFF module in integrating heterogeneous features. Compared to simple feature concatenation, the introduction of the AFF module consistently yields a substantial overall performance improvement. Particularly in real image testing, the BRISQUE metric decreases substantially from $23.98$ to $12.70$, representing a significant qualitative improvement. This provides strong evidence that simple channel concatenation introduces redundant noise and feature interference when processing umbra and penumbra features. 
Furthermore, the AFF achieves precise feature alignment and selective fusion via a dynamic attention mechanism. As shown in Fig. \ref{fig:Ablation_SR}(e), this mechanism effectively handles complex ground objects and irregular edges, ensuring a high degree of balance between global color consistency and local texture clarity in the reconstructed imagery.

\section{Conclusion and Future Works}
In this paper, we propose AeroDeshadow, a unified framework coupling physical priors with deep learning for ASI shadow synthesis and removal. To overcome the paired data bottleneck, we develop PDSS-Net, which accurately simulates real-world illumination attenuation and spatial decay to construct a large-scale, physically faithful dataset (AeroDS). Building upon this, we propose PCDS-Net, a penumbra-aware cascaded architecture that explicitly decouples the umbra and penumbra regions. Extensive experiments indicate that our framework achieves SOTA performance against previous methods. By balancing local texture recovery and global color consistency, PCDS-Net effectively mitigates boundary artifacts and generalizes robustly to complex real-world scenes.
Future work will focus on integrating the proposed deshadowing framework into downstream ASI tasks (e.g., object detection and segmentation) to enhance automated geospatial interpretation. Additionally, extending the physical degradation models to accommodate complex atmospheric occlusions, such as cloud shadows, remains a promising direction.

\bibliographystyle{IEEEtran} 
\bibliography{AeroDeshadow}

\newpage

\vfill

\end{document}